\documentclass[letterpaper, 10 pt, conference]{ieeeconf}  

\IEEEoverridecommandlockouts                              

\usepackage{amsfonts}
\usepackage{amsmath,amssymb,amsfonts}
\usepackage{subcaption}
\usepackage{graphicx}
\usepackage[ruled,vlined,linesnumbered]{algorithm2e}
\usepackage{algorithmic}
\usepackage{url}
\usepackage{booktabs}
\usepackage{nopageno}

\newcommand{\re}[1]{(\ref{#1})}

\DeclareMathAlphabet{\mathbcal}{OMS}{cmsy}{b}{n}
\title{\LARGE \bf
GloCAL: Glocalized Curriculum-Aided Learning of Multiple Tasks with Application to Robotic Grasping
}

\author{Anil Kurkcu$^{*}$, Cihan Acar, Domenico Campolo and Keng Peng Tee
\thanks{A. Kurkcu, C. Acar and K.P. Tee are with Institute for Infocomm Research, A*STAR, Singapore.}%
\thanks{D. Campolo is with Department of Mechanical \& Aerospace Engineering,
        Nanyang Technological University, Singapore.}
\thanks{$^*$ Corresponding author. Email: ANIL004@e.ntu.edu.sg}
}

\begin{document}

\maketitle

\pagenumbering{arabic}
\thispagestyle{plain}
\pagestyle{plain}

\begin{abstract}

The domain of robotics is challenging to apply deep reinforcement learning due to the need for large amounts of data and for ensuring safety during learning. Curriculum learning has shown good performance in terms of sample-efficient deep learning. 
In this paper, we propose an algorithm (named \textit{GloCAL}) that creates a curriculum for an agent to learn multiple discrete tasks, based on clustering tasks according to their evaluation scores. 
From the highest-performing cluster, a global task representative of the cluster is identified for learning a global policy that transfers to subsequently formed new clusters, while remaining tasks in the cluster are learnt as local policies.
The efficacy and efficiency of our \textit{GloCAL} algorithm are compared with other approaches in the domain of grasp learning for 49 objects with varied object complexity and grasp difficulty from the EGAD! dataset. The results show that \textit{GloCAL} is able to learn to grasp 100\% of the objects, whereas other approaches achieve at most 86\% despite being given 1.5$\times$ longer training time. 

\end{abstract}
\section{Introduction}

Recent years have showcased vast improvements in deep reinforcement learning applications, ranging from self-driving cars \cite{Makantasis_2020} to game play exceeding human performance \cite{silver2017mastering}. Reinforcement learning algorithms can now succeed in high-dimensional state spaces and continuous action spaces \cite{lillicrap2019continuous} thanks to the approximation power of deep neural networks \cite{lecun2015deeplearning}. Deep learning requires collection of large amounts of data for training \cite{8944013}, which is highly time-consuming in robotic applications, both in simulation and the real world \cite{doi:10.1177/0278364917710318,pinto2015supersizing}. Additionally, safety from physical collisions needs to be ensured during learning \cite{hunt2020verifiably}. As a result, deep reinforcement learning algorithms are challenging to apply to robotics when learning from scratch.

\begin{figure}[t]
      \centering
      \includegraphics[width=0.48\textwidth]{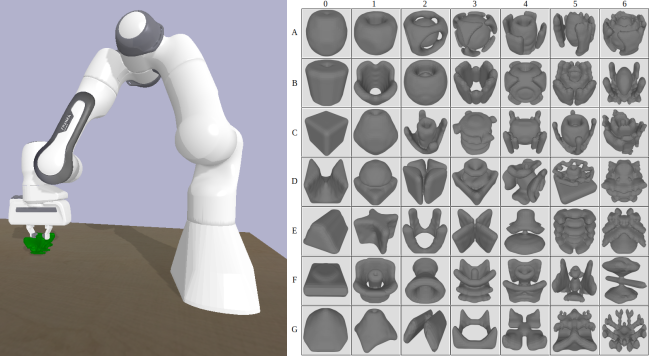}
      \caption{\small We showcase our algorithm in a robotic grasping environment setup, where the objects are from the EGAD! dataset \cite{morrison2020egad}. Our proposed algorithm GloCAL is able to learn grasping the whole set of objects, while other approaches including a learning-progress based method and a random curriculum were not able to do so.}
      \label{fig.obj_set}
\end{figure}

Sample-efficient learning strategies in the robotics literature include learning from demonstration \cite{argall2009survey} and domain randomization \cite{doi:10.1177/0278364919887447}. However, these methods require either human intervention or an (almost) perfect simulation environment, which constitute additional effort apart from the learning problem itself. Curriculum learning \cite{DBLP:conf/icml/HacohenW19} is a promising method for dealing with the problem of sample efficiency. Having its roots from developmental psychology \cite{elman1993learning}, this approach was initially used in deep learning where the goal was ordering training data based on a difficulty metric before feeding into a neural network \cite{bengio2009curriculum}. Curriculum learning is recently being applied with deep reinforcement learning on robotic applications involving tasks such as locomotion \cite{Florensa18}, \cite{Portelas19} and pick-and-place \cite{NEURIPS2020_566f0ea4}.

Although curriculum learning is suitable for learning in a sample-efficient manner, the issue is about generating the curriculum itself. Automatic curriculum learning \cite{ijcai2020-671}, \cite{kurkcu2020autonomous} is a recent field of research aiming at developing algorithms that could autonomously organize the ordering of tasks in multi-task problem setups. The benefit is a sample-efficient training routine that would have not been possible if the tasks were ordered randomly during training.

Regarding robotic grasping domain, grasp detection algorithms \cite{caldera2018review} based on deep learning approaches have been developed and tested on synthetic and real object datasets, achieving high success rates and reliable object grasping. Nonetheless, such approaches create a grasp pose without taking into consideration the closed-loop grasping behavior itself. On the other hand, reinforcement learning creates the whole trajectory of the grasping behavior, which is a more human-inspired way when teaching how to grasp an object \cite{FAGG1994281}. A curriculum-empowered deep reinforcement learning approach is necessary due to the fact that grasping environments are designed based on sparse and delayed rewards, saying that an object should be grasped and carried to a certain height before a reward signal is received \cite{9216986}. Such a reinforcement learning environment causes difficulty for agents that rely on pure deep reinforcement learning algorithms without any additional learning strategy \cite{NIPS2017_453fadbd}.

In this paper, we present an automatic curriculum learning algorithm that can design a curriculum for a learning agent to solve multiple tasks. We apply our approach on a set of grasping tasks with various difficulty levels and compare our results with other approaches including a learning progress-based algorithm and a random curriculum. Our proposed algorithm, \textit{\textbf{Glo}calized \textbf{C}urriculum-\textbf{A}ided \textbf{L}earning} or \textbf{\textit{GloCAL}} for short, generates a curriculum based on formation of task clusters according to the capabilities of an agent at a certain time instant. Inspired from the definition of glocalization \cite{roudometof2016glocalization}, our approach picks a global task from the cluster, updates its policy by training on that task and uses this updated policy to both learn the remaining tasks in the cluster named as local tasks and form new clusters. To the best of our knowledge, this is the first study applying curriculum learning to grasp a set of objects in a closed-loop manner. To summarize, our contributions in this paper are:
\begin{itemize}
\item An automatic curriculum learning algorithm via task clustering followed by global-local task decomposition for discrete set of tasks.
\item Application of GloCAL to robotic grasp learning of a set of objects with varied object complexity and grasp difficulty, and comparison of efficacy and efficiency with other approaches.
\end{itemize}

\section{Related Work}

\subsection{Automatic Curriculum Learning}

Curriculum learning was initially defined in \cite{elman1993learning}, \cite{bengio2009curriculum}, where training data was organized according to a difficulty metric. Once fed into a neural network, the ordered training data from easy to difficult resulted in shorter training times compared to a random ordering of training data. Empowered by such studies, curriculum learning has become a handy method for finding sample-efficient learning strategies \cite{pmlr-v70-graves17a}.

Although it is expected that a curriculum would speed up the learning of tasks by sequencing them from easy to hard, an issue that needs to be solved is how to estimate the difficulty level of a task. Despite the existence of approaches handcrafted for certain environments (e.g. grid-world domain \cite{narvekar2020curriculum}), we need more general ways of generating the curriculum itself, which is the main focus point for Automatic Curriculum Learning, defined \cite{Portelas20} as \textit{``approaches able to autonomously adapt their task sampling distribution to their evolving learner with minimal expert knowledge."}

An ACL approach based on a generative network for sampling medium-difficult tasks in a goal-based environment was proposed in \cite{Florensa18}. 
Their approach benefits from the idea of tackling tasks that could be learned by the agent based on its capabilities at that time instant, ignoring too simple or very difficult tasks. One downside of pursuing a generative model for an automatic curriculum lies in the architecture and training of the model itself which could not be that intuitive to decide upon.

Another approach for ACL was proposed in \cite{Portelas19} based on the notion of Absolute Learning Progress (ALP) and Gaussian Mixture Models (GMM) for tackling problems in continuously parameterized environments. The algorithm fits GMMs where the ALP signal is high, meaning that tasks within that region are suitable for the agent to learn. However, for discrete task environments considered in this paper, a continuous ALP signal may not be available, so the performance of the ALP approach in such environments warrants further study. 

The value disagreement approach is a recent goal sampling strategy for ACL \cite{NEURIPS2020_566f0ea4}. Based on the value function of the base reinforcement learning algorithm, empirical results showed that learning performance was improved with this approach compared to algorithms in \cite{Florensa18} and \cite{Portelas19}. Despite its success in goal-based problem setups (i.e. single task), it is not easy to apply to our problem which involves learning multiple tasks.

\begin{figure*}[ht]
\centerline{\includegraphics[width=1.0\textwidth]{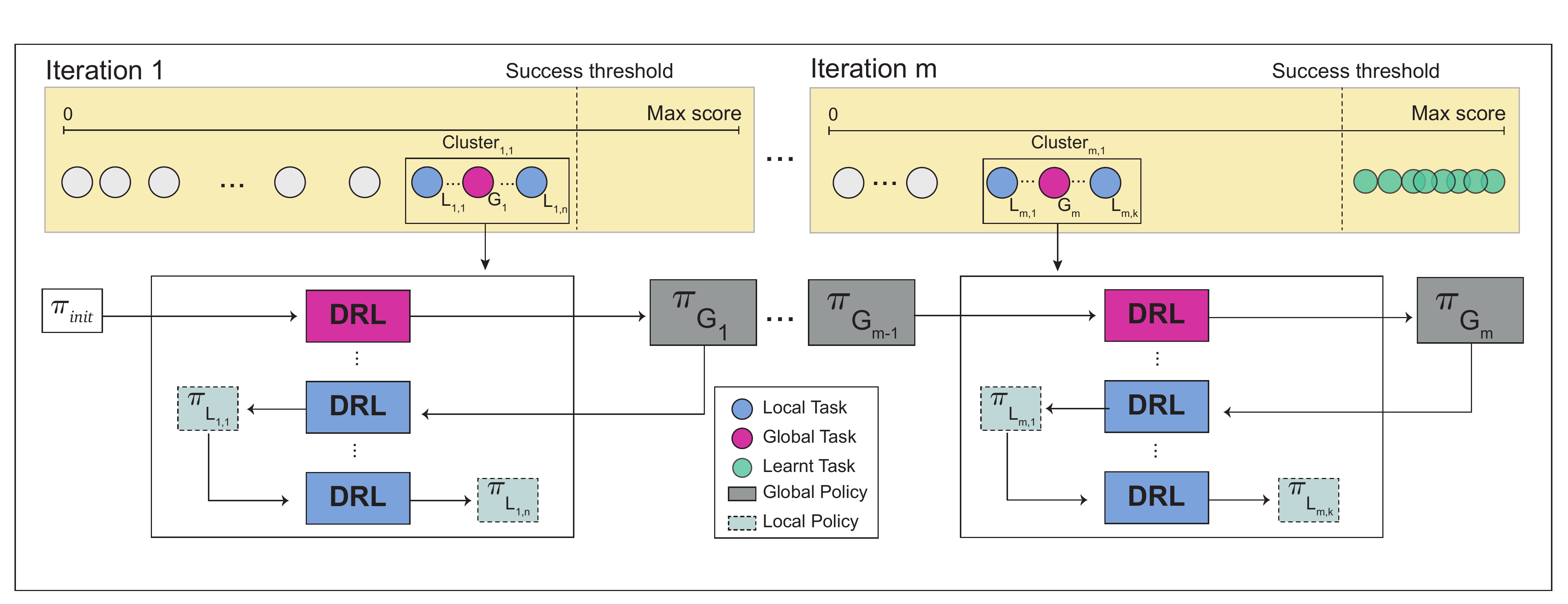}}
\caption{\small Overview of our algorithm. In the first iteration, a prior $\pi_{init}$ is used for clustering tasks. The first cluster is selected and tasks are marked as global and local. Training on the global task generates $\pi_{G_{1}}$, which is used to train the remaining local tasks. In iteration $m$, $\pi_{G_{m-1}}$ is used as a prior to cluster the remaining tasks. The result is an order of tasks for curriculum learning.}
\label{overview}
\end{figure*}

\subsection{Robotic Grasp Learning}

Deep learning-based grasping strategies are usually concerned with grasp pose detection \cite{inproceedingsfox,Murali20}, or use human demonstrations as a prior to speed up grasping algorithms \cite{Gabellieri20,Bonardi19}.

A comparison of various deep reinforcement learning algorithms applied to grasping was made in \cite{8461039}, where it was suggested that for the problem of grasping, off-policy reinforcement learning algorithms have a higher chance of success thanks to the replay buffer that the agent could make use of by sampling from its previous experiences in novel environments.

Grasp learning problems have been studied in pick-and-place problems involving a single cuboid object \cite{plappert2018multigoal}. Rather than focusing on the learning strategy for grasping, the work studied the performance of the algorithm for goal-based reinforcement learning environments, where the object is picked and placed at varying locations. Our problem setup focuses directly on the learning of a grasping policy with objects of various geometries provided as different tasks to the agent. These objects are spawned in random orientations for assessing the generalization capabilities of the agent.


\section{Automatic Curriculum Learning}

Our approach takes into consideration reinforcement learning environment setups which are based on a discrete set of tasks. We would like to highlight that in this paper, a task refers to the problem of grasping an object. With this in mind, we come up with a strategy that attempts to learn a number of tasks to a certain success rate threshold. Based on the capability of the learning agent, tasks are clustered according to their success rates obtained during evaluation. The cluster with the highest average score is chosen and a representative task is determined within this cluster. The agent first attempts this task, which we name as the \emph{global task}, and updates its policy ( i.e. global policy). The remaining tasks in the cluster, which we denote as \emph{local tasks}, are then tackled by the agent. However, local tasks do not update the global policy that will be used in the learning of tasks in subsequently formed clusters. As a result, a sequence of tasks are formed that the agent has followed as a curriculum where multiple policies exist for each learned task. An overview of our algorithm is given in Figure \ref{overview} and the pseudocode \textbf{Algorithm 1}.

\subsection{Initialization}

Let each task be given a number $n$ where $n=1,2,...,N$, and the full set of tasks represented by $\mathbf{N}$, where $|\mathbf{N}|=N$. Let $\pi_{init}$ be an initial/prior policy used to initialize the algorithm.

The set $\mathbf{L^{learnt}}$, to contain learnt tasks, is initialized as an empty set. The constant number $B$ is a success threshold such that any task with evaluation score above $B$ is marked as learned and stored in $\mathbf{L^{learnt}}$. Then, the remaining set of tasks after removing $\mathbf{L^{learnt}}$ is given by:
\begin{equation}
	\mathbf{L} = \mathbf{N} \setminus \mathbf{L^{learnt}}
	\label{rem_task}
\end{equation}

\subsection{Evaluation}

We define a score metric $S_{n}^{\pi}$ to assess the success rate of a policy on a certain task as follows

\begin{equation}
	S_{n}^{\pi} = \sum_{i=1}^{\Bar{S}} S_{i,n}^{\pi}, \quad n=1,2,...,N
	\label{eval}
\end{equation}
where $S_{i,n}^{\pi}\in[0,1]$ is an evaluation score of policy $\pi$ on task $n$ for the $i$th evaluation trial with $\Bar{S}$ being the number of times this evaluation is performed, $1$ being success and $0$ being failure.

\begin{algorithm*}
{\small
\DontPrintSemicolon
	\SetAlgoLined
	\textbf{Data:} Set of tasks $\mathbf{N}$, $|\mathbf{N}|=N$ where $n=1,2,...,N$\;
	\textbf{Input:} Initial policy $\pi_{init}$\;
	\textbf{Initialize:} Success threshold $B$, empty list of learnt tasks $\mathbf{L^{learnt}}$, $\pi_{curr}\gets \pi_{init}$\;
	\For{$m=1,2,...,$ max\_iter}{
	    Remaining tasks $\mathbf{L} = \mathbf{N} \setminus \mathbf{L^{learnt}}$\;
		Evaluate $\pi_{curr}$ on all $\mathbf{L}$ tasks, obtain $S_{n}^{\pi}$, $n=1,...,N$\;
		Apply $k$-means clustering on $\mathbf{L}$, determine number of clusters $c$ according to silhouette coefficient\;
		Obtain cluster set $\mathbf{C} = \{\mathbf{C_{1}}, \mathbf{C_{2}},...,\mathbf{C_{c}}|\mathbf{\Bar{C}_{1}}>\mathbf{\Bar{C}_{2}}>...>\mathbf{\Bar{C}_{c}}\}$\;
		Pick cluster $\mathbf{C_{1}}$ which has highest average task score \;
		Set median of $\mathbf{C_{1}}$ as global task $\mathcal{T}_{glob}$\;
		Label remaining tasks in $\mathbf{C_{1}}$ as local task set $\mathbcal{T}_{loc} = \mathbf{C_{1}} \setminus \mathcal{T}_{glob}$\;
		Train $\pi_{curr}$ on $\mathcal{T}_{glob}$ until $S_{\mathcal{T}_{glob}}^{\pi_{curr}}>B$, \quad $\pi_{G_{m}} \gets \pi_{curr}$\;
		Push $\mathcal{T}_{glob}$ to $\mathbf{L^{learnt}}$\;
		Set global policy as prior for learning local task $\mathcal{T}_{loc,0}$, $\pi_{L_{m,-1}} \gets \pi_{G_{m}}$\;
		\For{$i=0,...,|\mathbf{C_{1}}|-1$}{
			        Train $\pi_{L_{m,i-1}}$ on $\mathcal{T}_{loc,i}$ until $S_{\mathcal{T}_{loc,i}}^{\pi_{L_{m,i-1}}}>B$, \quad $\pi_{L_{m,i}} \gets \pi_{L_{m,i-1}}$\;
			        Push $\mathcal{T}_{loc,i}$ to $\mathbf{L^{learnt}}$\;
		}
		$\pi_{curr}\gets \pi_{G_{m}}$\;
	}
	\textbf{Result:} Sequence of tasks $\mathbf{L^{learnt}}$\;
	\caption{GloCAL: Glocalized Curriculum-Aided Learning}
	}
\end{algorithm*}

\subsection{Task Clustering}

At the start of each task clustering iteration, the evaluation of current policy $\pi_{curr}$ is performed with \re{eval} for each task $n=1,2,..N$. A score $S_{n}^{\pi}$ is obtained from each task for $\Bar{S}$ times. According to these score values, tasks are grouped via $k$-means clustering algorithm where the minimum and maximum number of clusters are empirically chosen. To determine the optimal number of clusters, we use silhouette analysis \cite{ROUSSEEUW198753}. For each cluster number, the silhouette score is computed as a value between -1 and 1. When the score is close to 1, it shows us that the clusters are well separated from each other. On the other hand, a score of 0 shows that there is no significant distance between clusters, and a score of -1 means that clusters are wrongly separated. Based on these score interpretations, the cluster number $c$ is selected according to the highest silhouette score. For the case when only 2 tasks are left, a single cluster is formed. The set of clusters are represented as $\mathbf{C}$ (where $|\mathbf{C}|=c$) as follows:
\begin{equation}\label{cluster}
	\mathbf{C} = \{\mathbf{C_{1}}, \mathbf{C_{2}},...,\mathbf{C_{c}}\in \mathbf{L}|\mathbf{\Bar{C}_{1}}>\mathbf{\Bar{C}_{2}}>...>\mathbf{\Bar{C}_{c}}\}
\end{equation}
where $\mathbf{\Bar{C}_{i}}$ is the average of the scores in cluster $\mathbf{{C}_{i}}$. In other words, the ordering of clusters in \re{cluster} are from highest to lowest average success rates, so $\mathbf{C_{1}}$ contains tasks with the highest success rate. Since tasks with higher success rate have greater chance of being learnt, the algorithm picks the first cluster $\mathbf{C_{1}}$, which is an ordered set, such that the evaluation score satisfy
\begin{equation}\label{cluster_1}
	S_{C_{1,1}}^{\pi}>...>S_{C_{1,i}}^{\pi}>...>S_{C_{1,n}}^{\pi}
\end{equation}
where $C_{1,i}$, $i=1,...,n$ are the individual tasks in cluster $\mathbf{C_{1}}$. From this set of tasks, the median is selected as the global task 
$\mathcal{T}_{glob}$ as follows:
\begin{equation}\label{global}
	\mathcal{T}_{glob} = \lceil \mathcal{M}(\mathbf{C_{1}}) \rceil
\end{equation}
where $\mathcal{M}(\bullet)$ denotes the median of set $\bullet$. According to \re{cluster_1}, tasks in $\mathbf{C_{1}}$ are in descending order, implying that global task $\mathcal{T}_{glob}$ has a score value in between the highest and lowest scores within the chosen cluster $\mathbf{C_{1}}$. We apply ceiling operator $\lceil \rceil$ in case the number of tasks is even.

Once the global task $\mathcal{T}_{glob}$ is determined, the remaining tasks in $\mathbf{C_{1}}$, after removing $\mathcal{T}_{glob}$, are labelled as local tasks $\mathbcal{T}_{loc}$ defined by:
\begin{equation}\label{local}
	\mathbcal{T}_{loc} = \mathbf{C_{1}} \setminus \mathcal{T}_{glob}
\end{equation}

\subsection{Curriculum Learning}

After tasks in $\mathbf{C_{1}}$ are marked, a two-level training regime is employed. The policy $\pi_{curr}$ is trained on global task $\mathcal{T}_{glob}$ until the evaluation score $S_{\mathcal{T}_{glob}}^{\pi_{curr}}$ surpasses threshold $B$. The policy updated at the end of this training session is named as $\pi_{G_{m}}$ (where $m$ is the iteration number) and global task $\mathcal{T}_{glob}$ is added to the set of learnt tasks $\mathbf{L^{learnt}}$. The updated policy $\pi_{G_{m}}$ is then used to train local tasks in $\mathbcal{T}_{loc}$. Task $\mathcal{T}_{loc,0}$ is trained first until the threshold $B$ is surpassed, i.e. $S_{\mathcal{T}_{loc,i}}^{\pi_{L_{m,i-1}}}>B$. The policy updated at the end of this training session is denoted by $\pi_{L_{m,0}}$, and $\mathcal{T}_{loc,0}$ is added to the set of learnt tasks $\mathbf{L^{learnt}}$. The updated policy $\pi_{L_{m,0}}$ is then used to train the next local task $\mathcal{T}_{loc,1}$. This loop continues until task $\mathcal{T}_{loc,|\mathbf{C_{1}}|-1}$ achieves an evaluation score above threshold $B$. When training both the global and local tasks, there exists a maximum amount of training time, after which, if the success threshold is still not surpassed, then training is stopped and the unsuccessful task is passed over to the next iteration.

\begin{figure*}[h]
\centerline{\includegraphics[width=1.0\textwidth]{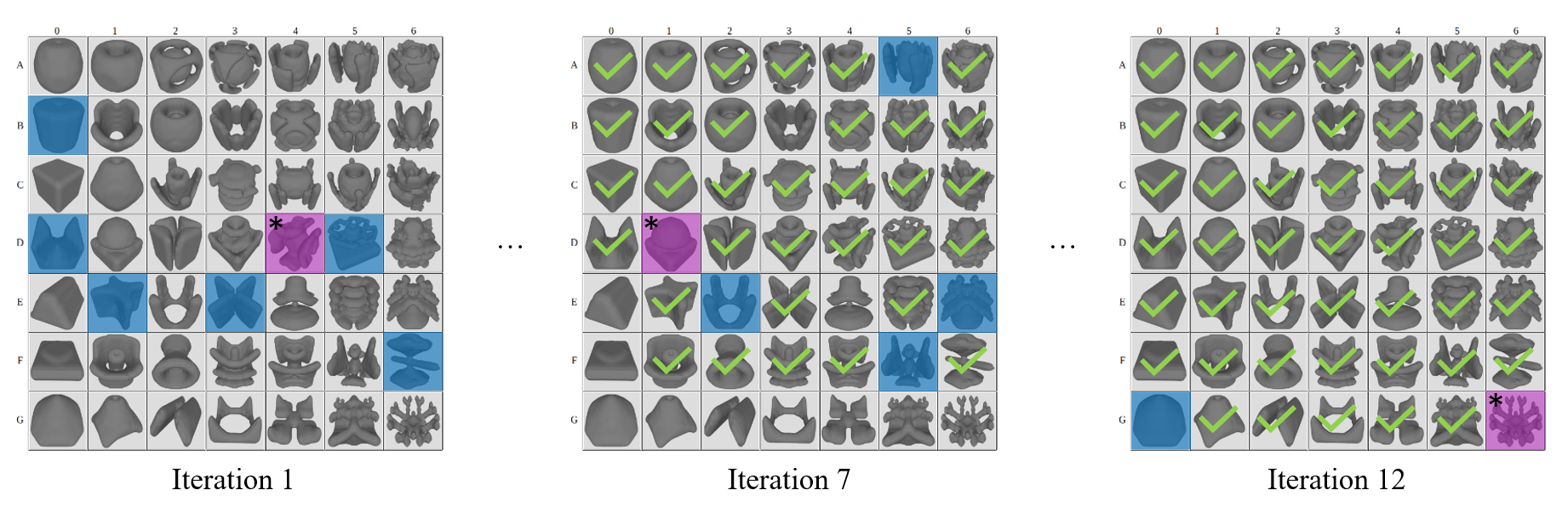}}
\caption{\small Sample run of our algorithm with beginning of first iteration 1, middle iteration 7 and last iteration 12 where all tasks are learned at the end. Objects highlighted pink with asterisk (*) are global tasks, whereas those highlighted blue are local tasks. Objects marked with a green tick mark are the ones that have been learned.}
\label{ours}
\end{figure*}

At the end of iteration $m$, all learnt tasks in $\mathbf{C_{1}}$ have entered $\mathbf{L^{learnt}}$. The next iteration $m+1$ starts by setting $\pi_{G_{m}}$ as $\pi_{curr}$. Task clustering is performed again followed by curriculum learning. The algorithm runs for a number of iterations, or until all tasks have a score value above $B$. As a result, we have a sequence of tasks that represent a curriculum. A sample run on a set of objects with 12 iterations in total is presented in Figure \ref{ours}.

We explain the methodology behind such global/local training as follows. A policy following a curriculum could experience various difficulty levels based on the task setup. Because of this, one would like to train a policy on each difficulty level with the same amount of time. Our approach is taking this into consideration and not training the curriculum policy on tasks that are all on the same difficulty level, which could be related to overfitting.

\section{Experimental Setup \& Results}

\subsection{Robot Environment Setup}

We have used PyBullet \cite{coumans2013bullet} robot simulator to perform our experiments where the robot manipulator is a Panda Franka Emika that has 7 degrees of freedom and a gripper with two parallel fingers. To obtain a diverse set of objects for the application of grasping, we have used objects from the evaluation set of the EGAD! Dataset \cite{morrison2020egad} consisting of $N$=49 objects (also known as tasks). These objects were created in the aforementioned study based on metrics of object complexity and grasp difficulty, where within a \{A,B,...,G\}$\times$\{0,1,...,6\} matrix of objects (Figure \ref{fig.obj_set}), object A0 is the simplest and easiest to grasp, while object G6 is the most complex and hardest to grasp. 
The resulting discrete set of tasks is suitable for us to test our algorithm. Objects were spawned on a table top with randomized orientations during training and evaluation, which helps to increase grasp robustness after learning.  

The reinforcement learning environment is defined as a Markov Decision Process (MDP) tuple $\langle S,A,P,R,\gamma \rangle$ which consists of a 63-dimensional continuous state-space $S$, a 5-dimensional continuous action-space $A$, a sparse reward function $R$, state transition model $P=Pr(s^{'}|s,a)$ and discount factor $\gamma$. Episode length is denoted by $T$. Each object is represented as a one-hot vector of length $N$=49. The agent receives, as input, this encoding, in addition to gripper and object poses, and two binary variables for gripper finger contact. Although not considered in this simulation study, 6D object pose estimation methods such as \cite{wada2020morefusion}, \cite{Cheng19} can be used in a real-world setup. 
Outputs of the agent include opening/closing of fingers, gripper rotation about normal axis to table plane, and 3D translation of gripper. The agent receives a scalar reward of -1 at each timestep unless the episode is terminated, which occurs either when the maximum episode length is reached, or when the agent has succeeded. Success condition is based on the height difference between the object and the table top. We have used the Soft Actor-Critic (SAC) \cite{pmlr-v80-haarnoja18b}, a state-of-the-art off-policy maximum entropy deep reinforcement learning algorithm, as our algorithm of choice due to its good performance in sparse reward setups. In particular, we used the implementation from Stable Baselines \cite{stable-baselines} in this work.

Regarding the evaluation phase, we have not used the cumulative reward that an agent receives during each episode as this would be tricky for a grasping problem with sparse rewards. Instead, we evaluate our policy based on 100 runs of a specific object. Each successful grasp increases the score by +1 for the agent, so the maximum obtainable score for an agent during evaluation is 100. Where no successful grasps could be performed, the score is 0.

\begin{figure*}[htbp]
\centerline{\includegraphics[width=0.70\textwidth]{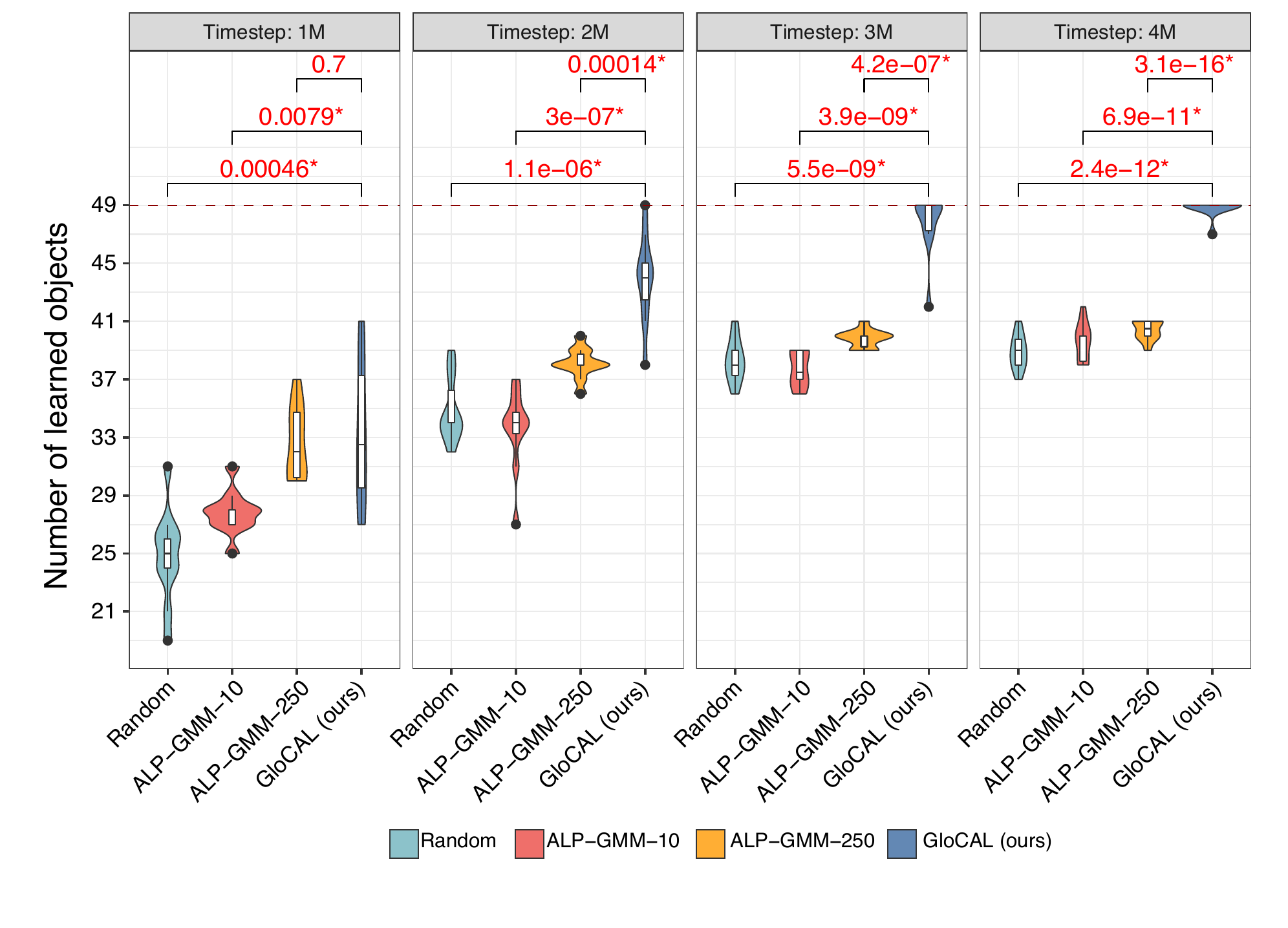}}
\caption{\small Comparison of all four algorithms at 1M, 2M, 3M and 4M timesteps respectively, from left to right. 10 random seeds have been performed for each algorithm. The p values of t-tests are provided for our algorithm versus others with significant values marked with asterisk (*). At the end of 1M timesteps, ours does not perform better than ALP-GMM-250. At the end of 2M timesteps, ours performs better than the rest. At the end of 3M timesteps, ours has some seeds that have reached the 49 out of 49 learned object ratio, while other 3 algorithms could not pass 42 objects. At the end of 4M timesteps, ours has mostly completed learning all 49 objects, while the other 3 approaches are still struggling around 40 learned objects. Though not presented here, at the end of 5M timesteps, all 10 runs of our algorithm has been able to grasp all the objects in the dataset, whereas the same is not true for the other three approaches.}
\label{violin}
\end{figure*}

\subsection{Benchmarking Algorithms}


\subsubsection{Random Curriculum}

This simple algorithm generates a curriculum by randomly selecting objects for training. A single policy is updated during training. Such an approach may be adequate if all objects are of the same difficulty level in terms of grasping. Usage of this curriculum generation method can also be seen as an ablation study for understanding whether there is any need for ordering objects based on their difficulty levels.

\subsubsection{ALP-GMM}
The Absolute Learning Progress-Gaussian Mixture Models (ALP-GMM) \cite{Portelas19} method is a state-of-the-art curriculum generation algorithm that works in both goal-based environments and continuous parameter spaces. This approach relies on the absolute learning progress, which is the difference between the reward values received of two consequent training intervals. The idea is to sample tasks that have the highest absolute learning progress, based on the idea that these tasks are the most learnable ones for a policy. The sampling is performed based on a Gaussian distribution where a new distribution is fit to the task space after a number of iterations each time. With some minor adaptations, we applied ALP-GMM to the discrete task problem in this paper. We compared with 2 versions of ALP-GMM: the original version which updates distributions every 250 iterations (ALP-GMM-250), and an alternate version which updates distributions every 10 iterations (ALP-GMM-10) for potentially better performance in a discrete task space. While the original implementation \cite{Portelas19} used a single policy throughout all iterations, the implementation in this paper is such that multiple policies achieving above-threshold success rates are saved, allowing for a fair comparison with our proposed GloCAL.

\begin{figure}[htbp]
\centerline{\includegraphics[width=0.50\textwidth]{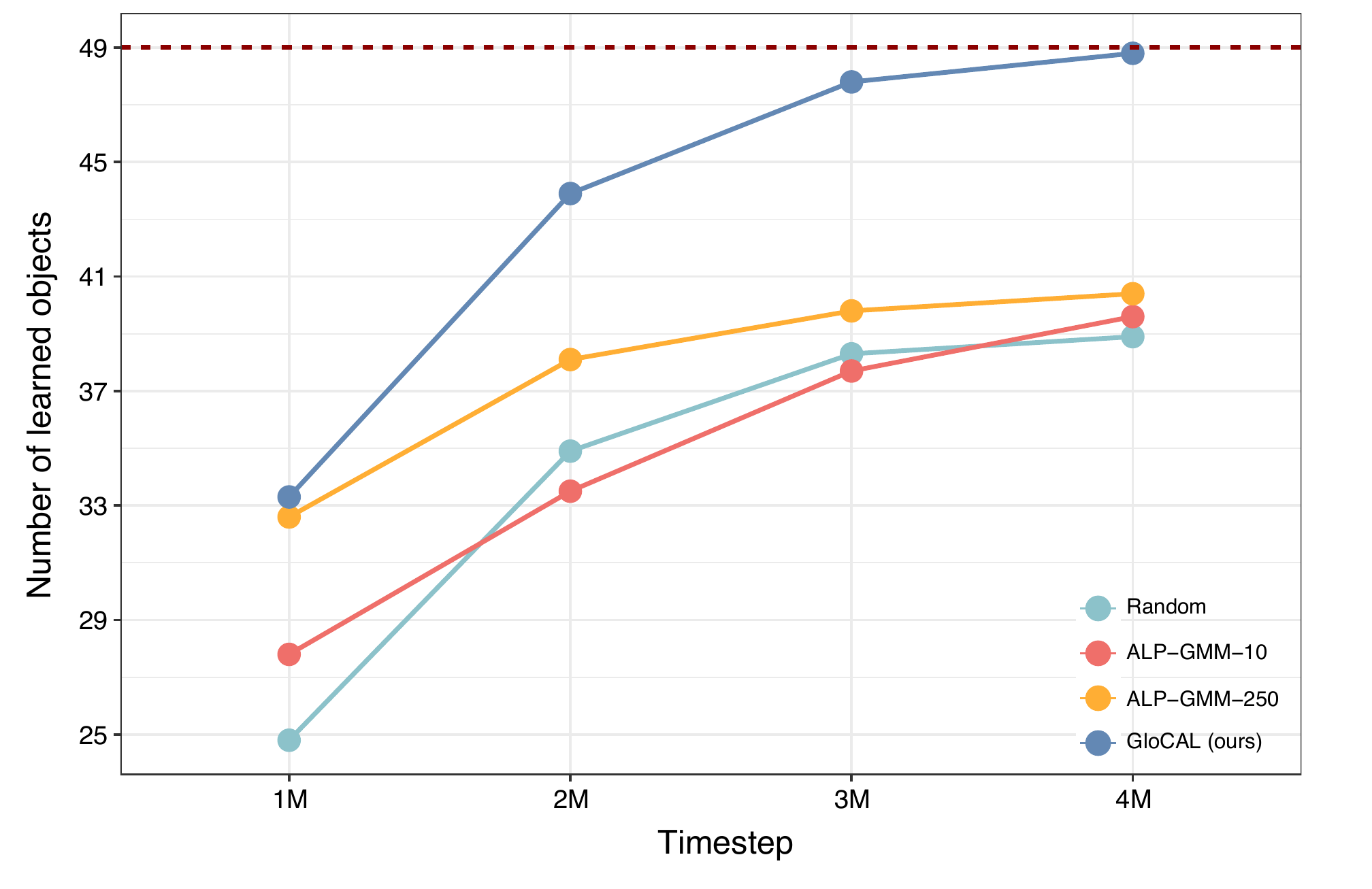}}
\caption{\small Comparison of number of objects learned for all four algorithms with recorded values at every 1M timesteps across 10 random seeds. All three algorithms apart from ours converge towards 40 objects, whereas ours converges towards all 49 objects.}
\label{line}
\end{figure}

\subsection{Results}

Here we present a quantitative comparison of our algorithm with the other two curriculum strategies in Figure \ref{violin}. The aim of each individual curriculum approach is to leverage the success rate of each object during evaluation above a certain value. Our current results have set this value to be 75 out of the 100 evaluation runs performed for any object. Each algorithm is run for 10 random seeds until 4 million (4M) timesteps, where the number of objects learned is recorded every 1 millionth (1M) timestep. Each algorithm is given the same initial policy which is a prior trained on object A0. We have adapted such a prior policy approach to only focus on the task ordering performance of the algorithms and eliminate the random effect caused by different starting conditions and alter the long training regimes in sparse reward setups including reaching object before grasping.

\begin{figure*}[ht]
\centerline{\includegraphics[width=1.0\textwidth]{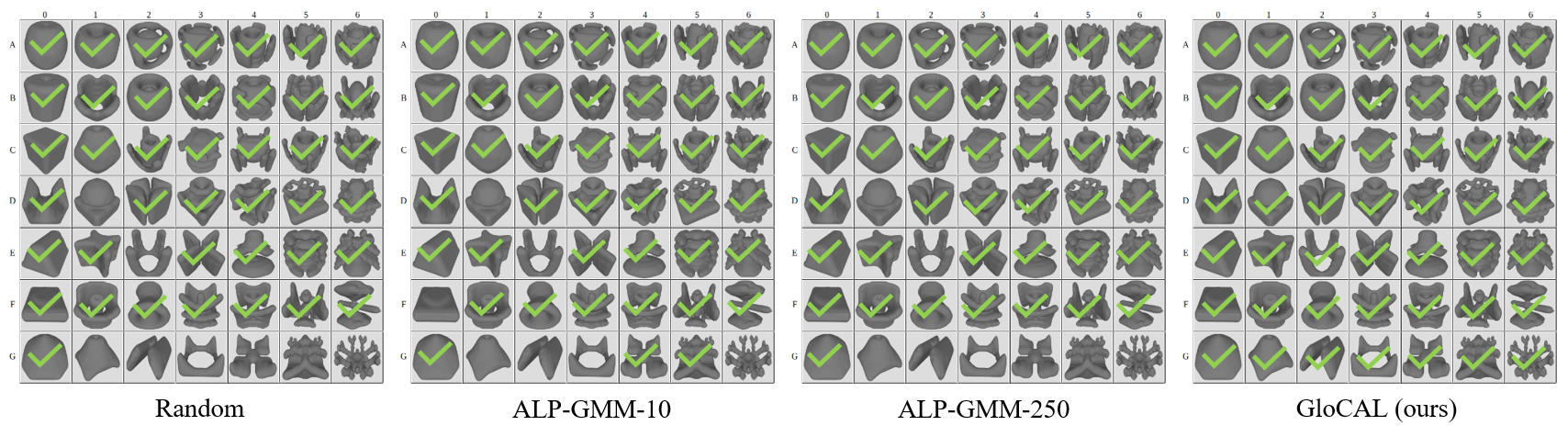}}
\caption{\small Comparison of objects learned at the end of 4M timesteps. Random and ALP-GMM-250 both learned 41 objects (83.6\%), ALP-GMM-10 42 objects ( 85.7\%), and our {GloCAL} learned all 49 objects (100\%). Objects marked with the letter G (last row) are the ones that are most difficult to grasp with respect to the difficulty metric provided in EGAD! \cite{morrison2020egad}.}
\label{obj_comp}
\end{figure*}

As it can be seen in Figure \ref{line}, our algorithm is the only one that can learn to grasp all 49 objects. Although the number of objects learnt at the end of 1M timesteps is similar between our algorithm and ALP-GMM, this changed from 1M to 4M timesteps, where our algorithm learned more objects compared to the rest. At the end of 4M timesteps, number of learnt objects for our algorithm converged to 49, whereas that for the rest failed to exceed 42. Though not presented in Figure \ref{line}, these numbers remained the same even if we waited until 6M timesteps. This shows our algorithm is more sample efficient than the rest and is able to learn to grasp the most difficult objects, with objects highlighted in Figure \ref{obj_comp}.

\subsection{Ablation Studies}

\begin{figure}[htbp]
\centerline{\includegraphics[width=0.45\textwidth]{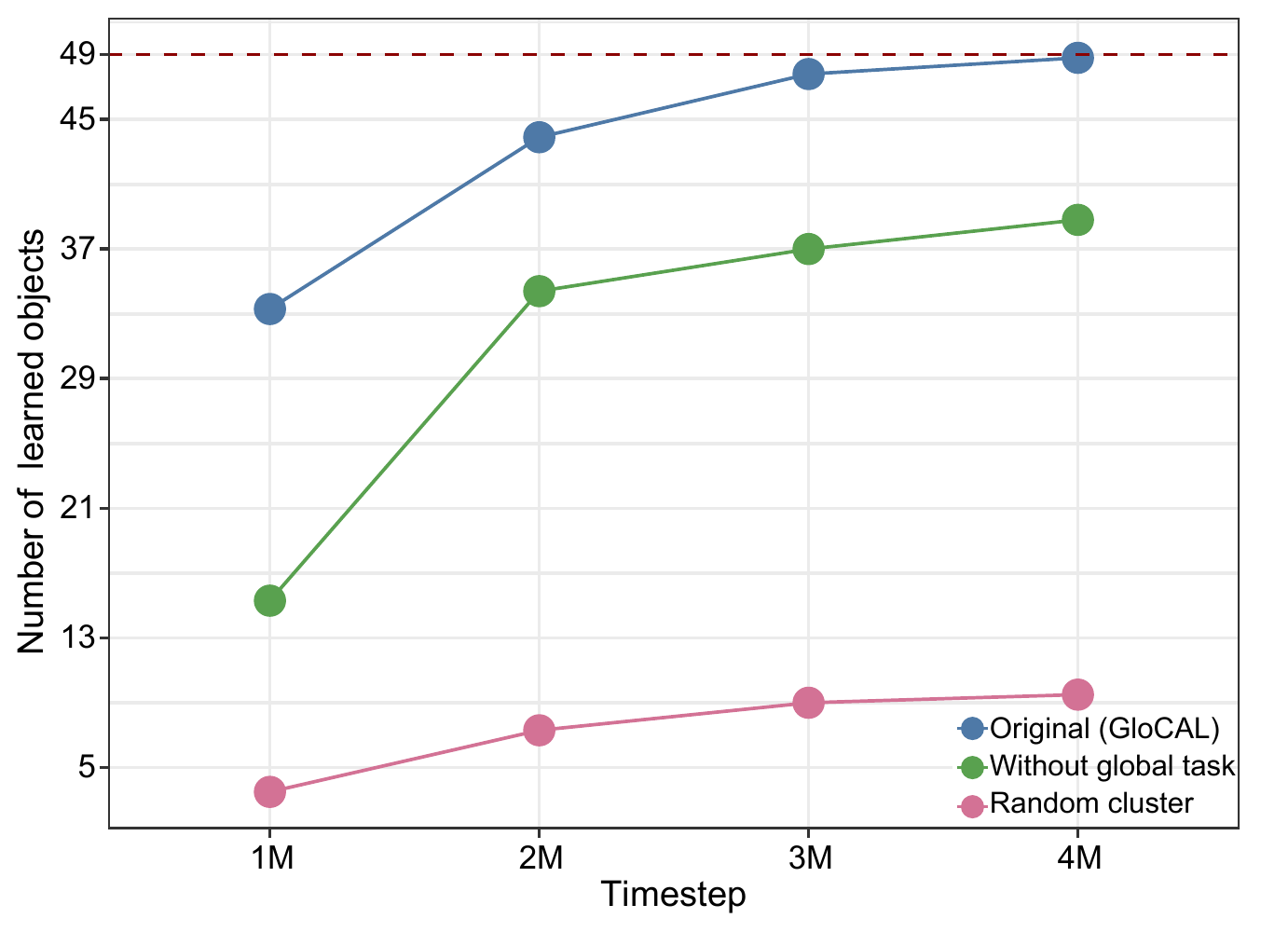}}
\caption{\small Comparison of number of objects learned for both the original and two ablated versions with recorded values at every 1M timesteps across 10 random seeds. Only the original version converges towards all 49 objects.}
\label{line_ab}
\end{figure}

We have performed two ablations to GloCAL, with one being the removal of the global representative task, and the other being the random selection of a cluster. As seen in Figure \ref{line_ab}, if we use a policy  trained over all tasks within the cluster (instead of a global representative task) as a prior for the next iteration, this results in only 39/49 objects being learned at the end of 4M timesteps. Even worse, instead of choosing the cluster with the highest average score, randomly selecting a cluster from the available ones results in only 10 objects being learned at the end of 4M timesteps.

This ablation study suggests that picking the cluster with the highest average score (instead of a random one) plays a critical role in successfully learning all tasks. Since there may exist a number of clusters containing difficult tasks, randomly picking a cluster could result in starting with difficult tasks that cannot be learned without a good prior policy. In addition, when moving from the current cluster to a more difficult one, if we do not pass a representative policy for the current cluster, but instead one that arbitrarily sweeps through all tasks within the cluster, this results in difficult tasks not being learned. The reason for this is that once a policy gets trained with too many tasks of a certain difficulty level, it becomes less adaptive to tasks that belong to another difficulty level, and sometimes may not learn those tasks at all.

\section{Discussion \& Conclusion}

The implication of our results says this: train a policy first on the task that represents a certain difficulty level, followed by utilizing the policy both to learn the tasks remaining in that difficulty level and as a prior for the remaining tasks. Once a policy is obtained for a certain difficulty level, this policy could be used to learn novel tasks of the same difficulty level, leading a path towards meta learning \cite{Lake1332}: when a task of similar representation to the tasks in a certain difficulty level is obtained, this new task can quickly be learned with the policy obtained from training on the global task. In addition to this, a parallel learning approach could also be employed with our algorithm for the purpose of speeding up runtime. Once a policy is obtained from the global task, the local training regime could be separated from the global training regime.

We have showcased an automatic curriculum learning algorithm that clusters tasks and builds a two-stage curriculum based on global and local tasks. Our problem setup consisted of a reinforcement learning environment where a robotic agent learned to grasp a diverse set of objects. In addition to our approach, we also implemented ALP-GMM which is a learning-progress based approach and also a random curriculum approach for a benchmark. Comparison of our algorithm with the mentioned approaches showed that, although our approach was able to grasp all of the objects, the other two algorithms failed to learn grasping difficult objects. Therefore, they could only grasp around 40 of the objects out of the 49. These results show the power of our algorithm for problem setups with discrete set of tasks. As future work, we would like to extend our algorithm towards one that is capable of parallel processing; once a cluster is assigned its global and local tasks, the global policy could carry onward forming new clusters, while the training of the local tasks could be performed at the same time. We would also like to approach our algorithm from an adaptive policy perspective where the global task acts as a meta task, suiting well for few-shot learning on novel tasks that are similar to the global task which could be thought as unknown objects not existing in the dataset. Currently, we are working on testing the efficiency of our algorithm on a real-world experimental setup, where pose estimation and variational autoencoder (VAE) methods are utilized for detecting the pose and physical geometry of the objects, which also makes it possible to consider grasping unknown objects.




 \section*{ACKNOWLEDGMENT}
This research is supported by grant no. A19E4a0101 from the Singapore Government’s Research, Innovation and Enterprise 2020 plan (Advanced Manufacturing and Engineering domain) and administered by the Agency for Science, Technology and Research.

\bibliographystyle{ieeetr}
\bibliography{root}

\end{document}